\documentclass[journal]{vgtc}                
\ifpdf
  \pdfoutput=1\relax                   
  \usepackage{graphicx}                
  \DeclareGraphicsExtensions{.pdf,.png,.jpg,.jpeg} 
\else
  \usepackage{graphicx}                
  \DeclareGraphicsExtensions{.eps}     
\fi%

\graphicspath{{figures/}{pictures/}{images/}{./}} 

\usepackage{microtype}                 
\PassOptionsToPackage{warn}{textcomp}  
\usepackage{textcomp}                  
\usepackage{mathptmx}                  
\usepackage{times}                     
\usepackage{cite}                      
\usepackage{tabu}                      
\usepackage{booktabs}                  

\usepackage{subfiles}
\usepackage{xspace}
\usepackage{enumitem}
\usepackage{hyperref}
\usepackage{amssymb}
\usepackage{xspace}
\usepackage{amsmath}
\usepackage[dvipsnames]{xcolor}

\def\LS{Label-Trustworthiness\xspace}
\def\LC{Label-Continuity\xspace}
\def\lc{Label-C\xspace}

\def\lsc{Label-T\&C\xspace}

\def\LT{Label-Trustworthiness\xspace}
\def\LC{Label-Continuity\xspace}
\def\lc{Label-C\xspace}
\def\lt{Label-T\xspace}
\def\ltc{Label-T\&C\xspace}
\def\CHb{$CH_{btwn}$\xspace}

\newcommand{\rev}[1]{{#1}}

\onlineid{0}

\vgtccategory{Data Transformation}
\vgtcpapertype{algorithm/technique}

\title{
\textit{Classes are not Clusters}: Improving Label-based Evaluation of Dimensionality Reduction
}

\author{Hyeon Jeon, Yun-Hsin Kuo, Micha\"el Aupetit, Kwan-Liu Ma, and Jinwook Seo}
\authorfooter{

\item Hyeon Jeon and Jinwook Seo are with Seoul National University. E-mail: hj@hcil.snu.ac.kr, jseo@snu.ac.kr
\item Yun-Hsin Kuo and Kwan-Liu Ma are with the University of California, Davis. E-mail: 
\{yskuo, klma\}@ucdavis.edu
\item Micha\"el Aupetit is with Qatar Computing Research Institute, Hamad Bin Khalifa University. 
    E-mail: maupetit@hbku.edu.qa
}

\shortauthortitle{Biv \MakeLowercase{\textit{et al.}}: Global Illumination for Fun and Profit}

\abstract{
A common way to evaluate the reliability of dimensionality reduction (DR) embeddings is to quantify how well labeled classes form compact, mutually separated clusters in the embeddings. 
This approach is based on the assumption that the classes stay as clear clusters in the original high-dimensional space.
 However, in reality, this assumption can be violated;
  a single class can be fragmented into multiple separated clusters, and multiple classes can be merged into a single cluster.
 We thus cannot always assure the credibility of the evaluation using class labels.
 In this paper, we introduce two novel quality measures---\textit{\LT} and \textit{\LC} (\ltc)---advancing the process of DR evaluation based on class labels. Instead of assuming that classes are well-clustered in the original space, \lsc work by (1) estimating the extent to which classes form clusters in the original and embedded spaces and (2) evaluating the difference between the two. A quantitative evaluation showed that \lsc outperform widely used DR evaluation measures (e.g., Trustworthiness and Continuity, Kullback-Leibler divergence) in terms of the accuracy in assessing how well DR embeddings preserve the cluster structure, and are also scalable.
 Moreover, we present case studies demonstrating that \ltc can be successfully used for revealing the intrinsic characteristics of DR techniques and their hyperparameters.
} 


\CCScatlist{ 
 \CCScat{K.6.1}{Management of Computing and Information Systems}%
{Project and People Management}{Life Cycle};
 \CCScat{K.7.m}{The Computing Profession}{Miscellaneous}{Ethics}
}

\vgtcinsertpkg

\begin{document}


\firstsection{Introduction \label{sec:intro}}
\maketitle

Dimensionality reduction (DR) is one of the most widely used tools in conducting the visual cluster analysis of high-dimensional data \cite{ jiazhi21tvcg, pezzotti16cgf, jeon22arxiv, kwon18tvcg, quadri21tvcg, quadri22tvcg}. Using DR for cluster analysis relies on the assumption that the cluster structure of the original data is accurately represented in the low-dimensional DR embeddings. However, DR inherently generates distortions, i.e., the original cluster structure is imprecisely represented in the embeddings \cite{aupetit07neurocomputing, aupetit14vast, heulot13vamp, lespinats07tnn, lespinats11cgf}. As distortions can make visual cluster analysis performed with DR unreliable \cite{jeon21tvcg, jeon22arxiv},  it is important to evaluate how well the original cluster structure is preserved in the DR embeddings \cite{jiazhi21tvcg, jeon21tvcg, martins14cg, joia11lamp}, prior to the analysis. There exist ways to evaluate the reliability of cluster structures in DR embeddings, in either a perceptual \cite{etemadpour15ivapp, jiazhi21tvcg, sedlmair13tvcg} or computational \cite{jeon21tvcg, venna06nn, lee07springer, motta15neurocomputing} manner.

A general process to evaluate the preservation of cluster structure in DR embeddings is to utilize class labels. 
This is done by assessing \textit{cluster-label matching} (CLM), that is, the extent to which class labels form clusters in the embedded space \cite{joia11lamp, loch15neurocomputing, xia22tvcg, becht19nature, xiang21fig, yang21cellreports}.
CLM is mostly evaluated by using \textit{clustering validation measures} (CVMs) \cite{liu10icdm, wu09kdd}, such as the Silhouette Coefficient \cite{rousseuw87silhouette}.
CVMs inform how well the groups in the given label-based data partition form clear position-based clusters.
The partitions that contain mutually separated and individually condensed groups are preferred. 
For the label-based evaluation of DR, data embeddings and class labels are used as an input dataset and partition, respectively.
Embeddings with good CLM are considered to have good quality, assuming that the original data also have good CLM.

However, such an assumption can hardly be guaranteed \cite{aupetit14beliv, jiazhi21tvcg, farber10multiclust, jeon22arxiv2}.
\rev{There is no constraint on labels' sources. Labels can come from} an external source (e.g., human annotation), \rev{possibly unrelated to the features of the data space. Labels can also result from clustering techniques, which may not align with the actual clusters. 
Therefore, we do not know how well labels make up the clusters in the original data};
a single class can consist of multiple separated clusters, and multiple classes can be in close proximity or even overlapped \cite{aupetit14beliv} in a single cluster. 
These possibilities cast doubt on the conclusions derived from the general process of  label-based DR evaluation. 
For instance, an embedding that accurately represents overlapping classes in the original space might be considered to have low quality as it has bad CLM.

In this work, we revisit the process of evaluating DR using class labels. 
We introduce two measures---\textit{\LT} (\lt) and \textit{\LC} (\lc)---which examine CLM in an alternative way to assess the reliability of cluster structures in DR embeddings.
In contrast to the general label-based evaluation process, \ltc use CVM to quantify CLM \textit{distortions} as the difference between CLM estimated in both original and embedded spaces. 
\lt quantifies the distortion due to the degradation of CLM: the score is lower when the points of two different classes get closer in the embedding than in the original space. 
Conversely, \lc evaluates the distortion regarding the exaggeration of CLM: the score is lower when the points of two different classes get farther apart in the embedding than in the original space.
The rationale behind our measures is that in visual cluster analysis, it is important to investigate how class labels span the original cluster structure as seen through the embedding \cite{brehmer14beliv, aupetit14beliv, aupetit22arxiv, aupetit22topoinvis, wenskovitch18tvcg} (e.g., examine the individual density of a class or the pairwise proximity between classes). Since CLM distortions reduce the reliability of cluster structures represented by the embeddings, \ltc scores can be interpreted as proxies for the credibility of DR-based cluster analysis.

We conduct a series of quantitative experiments to validate the effectiveness of \ltc. 
The results show that \ltc can better capture the distortions of cluster structures than the existing measures (e.g., Steadiness \& Cohesiveness \cite{jeon21tvcg} and Trustworthiness  \& Continuity \cite{venna06nn}) and the general process of label-based DR evaluation (i.e., naive application of CVMs).
From the scalability analysis, we validate that the runtime of using \ltc is competitive with that of the existing methods.
Finally, we demonstrate two case studies showing that  \ltc can be used to reveal how different DR techniques or hyperparameter settings affect embedding results.

\section{Background and Related Works}

We discuss the state-of-the-art in interpreting and measuring the reliability of DR embeddings. 
We then describe works about the common assumption that high-dimensional labeled data have good CLM.

\subsection{Reliability of Dimensionality Reduction}

\rev{
\subsubsection{Dimensionality Reduction}
}

Dimensionality reduction (DR), e.g., $t$-SNE \cite{maaten08jmlr}, UMAP \cite{mcinnes2020arxiv}, aims to produce the low-dimensional embedding preserving the structure of the input high-dimensional data. \rev{DR plays an important role in many visual analytics tasks, including cluster identification \cite{jeon22arxiv, xia22tvcg} or neighborhood search \cite{etemadpour15ivapp, etemadpour15tvcg, lespinats11cgf}. This research provides reliable measures for evaluating DR embeddings regarding the matching between clusters and classes in both input and embedding spaces, establishing a basis for more trustworthy DR-based visual analysis. }

\rev{
\subsubsection{Distortions in Dimensionality Reduction}
}

While transferring the data from broad high-dimensional space to narrow low-dimensional space, DR unavoidably introduces distortions \cite{nonato19tvcg, aupetit07neurocomputing}.
As distortions make embeddings less reliable in representing the original data, informing distortions is important in utilizing DR for data analysis \cite{nonato19tvcg, jeon21tvcg}.

Several distortion types were defined to formally explain DR distortions. Aupetit \cite{aupetit07neurocomputing} initially defined \textit{stretching} and \textit{compression}. Stretching describes the situation in which the pairwise distances in the embedded space are increased compared to the ones of the original space; conversely, compression indicates the case that the pairwise distances decreased. \textit{Missing Neighbors} and \textit{False Neighbors} \cite{lespinats07tnn, lespinats11cgf, venna10jmlr, lee07springer} were introduced as an interpretation of stretching and compression in terms of the neighborhood structure. Given a high-dimensional point $x$ and its corresponding low-dimensional point $z$, Missing Neighbors are defined as the $k$-nearest neighbors of $x$ that are not among the ones of $z$. Conversely, False Neighbors are defined as the $k$-nearest neighbors of $z$ that are not among the ones of $x$. 
However, Missing and False Neighbors are insufficient to explain the distortions of complex, intertwined cluster structures. 
For example,  the relative increase of cluster density in the embedding does not incur Missing and False Neighbors distortions, because it does not alter the $k$-nearest neighbor structure for small $k$ values.

As alternatives, \textit{cluster-level} distortions, named \textit{Missing Groups} and \textit{False Groups}, were proposed by Jeon et al. \cite{jeon21tvcg}. Missing Groups  occur when a cluster in the original space splits into multiple separated clusters in the embedding, and False Groups occur when a cluster in the embedding consists of multiple separated clusters in the original space. 
In the seminal work \cite{jeon21tvcg}, Missing and False Groups distortions are examined based on the groups obtained by clustering techniques.

In this work, we focus on evaluating the reliability of the cluster structure of DR embeddings by quantifying both Missing and False Groups.
However, instead of extracting groups using clustering techniques, we focus on the groups given by the classes of labeled data.

\subsubsection{Distortion Measurement without Labels}

\label{sec:disme}

We discuss distortion measures that do not rely on class labels. 
These measures take the original and embedded data as input and quantify their structural difference.
Aligned with the aforementioned distortion types, they focus on three different levels of structural granularity: \textit{global}, \textit{local}, and \textit{cluster}.
\textit{Global measures}, such as Kullback-Liebler divergence (KL Divergence) and Distance to Measure (DTM) \cite{chazal11fcm, chazal17jmlr}, quantify how well the embeddings preserve the global structure of the original data against stretching and compression.
Meanwhile, \textit{local measures} focus on neighborhood preservation.
Trustworthiness and Continuity (T\&C) \cite{venna06nn} measure how Missing and False Neighbors affected the distance-based ranking of the nearest neighbor for every data point in both spaces. Mean Relative Rank Errors (MRREs) \cite{lee07springer} extends T\&C by additionally regarding the ranking of True Neighbors: the points that are neighbors in both the original and embedded spaces. 
Still, despite local and global measures' wide usage in practice \cite{jeon21tvcg, jeon22vis, lai22vis, xia22tvcg, nonato19tvcg, moor20icml}, they do not properly capture cluster-level distortions \cite{jeon21tvcg}. 

This leads to the necessity of measures that capture distortions on cluster structures (i.e., \textit{cluster-level measures}).
Steadiness and Cohesiveness (S\&C) \cite{jeon21tvcg}  assess how much Missing and False Groups  distortions have occurred by  (1) extracting clusters from one space and (2) evaluating their dispersion in the other space.
However, S\&C require users to specify the way of extracting and investigating clusters in both spaces, e.g., using clustering techniques, making the results of the cluster-level distortion measures sensitive to the clustering technique and hyperparameters used. 
S\&C also suffers from a scalability issue as it requires the iterative execution of a clustering technique \cite{jeon21tvcg, fujiwara22arxiv}.

\ltc is a pair of cluster-level measures that aim to tackle these limitations. At first, the measures require a CVM as the sole hyperparameter, which is used to evaluate CLM in the original and embedded spaces. 
Thanks to the low complexity of CVM \cite{liu10icdm, jeon22arxiv2}, our measures are very scalable (\autoref{sec:scaleval}). Furthermore, \lsc are more sensitive in distinguishing Missing and False Groups distortions compared to previous measures, including S\&C (\autoref{sec:seneval}).

\subsubsection{Distortion Measurement with Labels}

\label{sec:dislabel}

Exploiting labels is a common scheme in evaluating DR embeddings \cite{colange20neurips, joia11lamp, loch15neurocomputing, xia22tvcg, becht19nature, xiang21fig, yang21cellreports}.
A general process to do so is to utilize CVM to measure the CLM of embeddings \cite{joia11lamp, loch15neurocomputing, becht19nature, yang21cellreports}.
However, the approach is prone to producing errors while examining the quality of DR embedding.
For example, if the CLM of the original data is bad (e.g., some classes overlap), embeddings that have good CLM for bad reasons (e.g., DR artificially separates each class into a distinct cluster) will be considered high-quality embeddings. 
As non-expert users typically assume that DR techniques generate reliable embeddings of the original data, they may incorrectly conclude that CLM is also good in the high-dimensional space, while it is not actually true \cite{jeon22arxiv2, aupetit14beliv}.

A sole pair of measures that relies on class labels but is independent of CVM is Class-Aware Trustworthiness and Continuity (CA-T\&C) \cite{colange20neurips}.
CA-T\&C is a variant of T\&C that assess the degradation of CLM (i.e., False Groups distortions) by estimating 
the extent to which Missing and False Neighbors occurred within and between classes, respectively. 
However, CA-T\&C hardly captures the Missing Groups distortions as they do not consider the increase of CLM as distortions. 
The measures also mainly focus on local structures and thus cannot comprehensively examine CLM distortions.

In this work, we propose \ltc as novel measures utilizing class labels to evaluate  DR embeddings. 
As with the general process of label-based DR evaluation (i.e., the process of naively applying CVM in the embedded space), our measures utilize CVMs to evaluate CLM; however, by applying CVM to both the original and embedded spaces and assessing their difference, our measures precisely capture cluster-level distortions.

\subsection{The Cluster-Label Matching assumption}

The assumption that the CLM is good in the high-dimensional data is used as a basis not only for the label-based evaluation of DR embeddings but also for other applications. For example, the labels are often utilized as the ground truth partition in clustering validations, where clustering techniques that generate a similar partition to that of labels obtain higher scores (i.e., external clustering validation; refer to \autoref{sec:cvmdesc} for details). Another application is the perception-based evaluation of DR techniques \cite{xia22tvcg,etemadpour15tvcg, etemadpour15ivapp, sedlmair13tvcg}, where techniques that produce embeddings in which the visual clustering results of human subjects better match class labels are preferred. 
However, the assumption can be easily broken \cite{aupetit14beliv, farber10multiclust}, which casts doubt on the applications' reliability. 

Despite such a threat, only a few solutions have emerged. 
A trivial solution is to modify datasets to make them better satisfy the assumption. Aupetit \cite{aupetit14beliv, aupetit05neurocomputing} proposed to check the linear or nonlinear separability of classes and then merge overlapped classes or preserve one of them while removing the others \cite{aupetit14beliv}. However, classes can be separable but adjacent, not forming proper clusters (no low density or wide empty space between them).  Such a strategy also does not take into account whether each class forms a single, compact cluster. 
Another solution is to use synthetic datasets \cite{jeon21tvcg, moor20icml, jeon22vis}, where good CLM is guaranteed by design. Still, this makes the evaluation hardly generalizable to real data.
Alternatively, Jeon et al. \cite{jeon22arxiv2} suggested a systematic way to evaluate CLM; their purpose was to verify the validity of labeled datasets for use as clustering validation benchmarks.
Still, they suggested only utilizing datasets with good CLM, which reduces the number of available datasets for evaluating DR embeddings.

In this work, we neither verify the CLM of datasets in advance nor attempt to modify datasets to enhance CLM.
 Instead, we \textit{acknowledge} that datasets may not satisfy the CLM, and rather assess whether the degree of CLM, either high or low, in the original dataset is well preserved in the embedding.




\section{General Label-based DR Evaluation Process}

\label{sec:cvmproblem}

The general process of label-based DR evaluation mostly relies on CVMs.
We describe what CVMs are and the process of using them to evaluate CLM. We then discuss the pitfalls of the process. 


\noindent
\textbf{Notations}
We define a high-dimensional data $\mathbf{X} = \{\mathbf{x}_i \in \mathbb{R}^D, i = 1, 2, \cdots, N\}$.
We denote the low-dimensional embedding of $\mathbf{X}$ as $\mathbf{Z} = \{\mathbf{z}_i \in \mathbb{R}^d \mid i = 1, 2, \cdots, N\}$, where $D > d$. 
For any set $\mathbf{S}\in\{\mathbf{X},\mathbf{Z}\}$, the distance function $\delta$ satisfies $\delta(x,y) \geq 0$, $\delta(x, y) = \delta(y,x)$ and $\delta(x,y)=0$ if $x=y$  $\forall x, y \in \mathbf{S}$.
A partition of $\mathbf{S}$ is defined as $\mathbf{P}=\{P_1, P_2, \cdots, P_k\}$ satisfying $P_i\subseteq \mathbf{S}$, $P_i \cap P_j = \emptyset$ and $\cup^{k}_{i=1}P_i = \mathbf{S}$.
If a partition is defined by class labels, we denote the partition as $\mathbf{P}_L$.
A clustering technique $C$ takes $\mathbf{S}$ and $\delta$ as input and returns a partition $\mathbf{P}_C$ of $\mathbf{S}$. 


\subsection{Clustering Validation Measures}

\label{sec:cvmdesc}

Clustering validation measures (CVMs) evaluate how well-clustered the given partition (i.e., clustering) is in the given data. 
We use CVMs to find the optimal clustering technique or hyperparameter setting that produces the partition of the data that best matches its cluster structure.
CVMs are largely divided into two types: \textbf{internal CVM (IVM)}  \cite{liu10icdm, liu13tsmcb} and \textbf{external CVM (EVM)} \cite{wu09kdd}. 
IVMs evaluate a partition based on the internal structure of data.
Formally, the IVM score $m_{I}(\mathbf{P}, \mathbf{X}, \delta)$ quantifies how well the groups within the partition $\mathbf{P}$ of $\mathbf{X}$ are individually condensed and mutually separated in $\mathbf{X}$ based on  distance $\delta$. 
For example, the Silhouette Coefficient \cite{rousseuw87silhouette} examines how the within-group and between-group distances differ on average while using Euclidean distance as $\delta$. 
Alternatively, EVMs, such as the adjusted rand index \cite{vinh09icml}, rely on a ground truth partition $\mathbf{P}_{GT}$.
Here, the EVM score $m_{E}(\mathbf{P}, \mathbf{P}_{GT})$ simply quantifies the degree of matching between the given partition $\mathbf{P}$ and $\mathbf{P}_{GT}$, regardless of the internal cluster structure of $\mathbf{S}$. A higher score is assigned if $\mathbf{P}$ better matches with $\mathbf{P}_{GT}$. Data class labels $\mathbf{P}_L$ are typically used as ground truth $\mathbf{P}_{GT}$ \cite{farber10multiclust, jeon22arxiv2}.

\subsection{Using CVM to Evaluate CLM}

\label{sec:cvmprocess}

We use CVMs to quantify the CLM of a DR embedding as a proxy for its reliability \cite{joia11lamp, xia22tvcg, becht19nature, yang21cellreports}.
The process depends on the type of CVM:

\noindent
\textbf{IVM-based evaluation }
For a given embedding $\mathbf{Z}$, distance function $\delta$, and class labels $\mathbf{P}_L$, $m_I(\mathbf{P}_L, \mathbf{Z}, \delta)$ represents the CLM between 
 $\mathbf{P}_L$ and $\mathbf{Z}$. 
The Silhouette Coefficient is widely adopted in the visualization community \cite{wang18tvcg, joia11lamp, loch15neurocomputing, xia22tvcg, etemadpour15ivapp}. 
The Davies-Bouldin index \cite{davies79tpami} is preferable in the context of star coordinates and Radviz \cite{angelini22tvcg, caro10pakdd}.
Notably, while Distance Consistency (DSC) \cite{sips09cgf} was designed for DR visual quality evaluation \cite{espadoto21tvcg, sedlmair12cgf, sedlmair15cgf}, it can also be viewed as a CVM since it considers only the separation of class labels in the embeddings.

\noindent
\textbf{EVM-based evaluation  }
Given $\mathbf{Z}$, $\delta$, $\mathbf{P}_{L}$, and a clustering technique $C$ providing a partition $\mathbf{P}_C = C(\mathbf{Z}, \delta)$ of the embedded data,
$m_E(\mathbf{P}_{C}, \mathbf{P}_L)$  represents CLM between $\mathbf{P}_L$ and $\mathbf{Z}$.
$K$-Means and the adjusted rand index are commonly used for $C$ and $m_E$, respectively \cite{zubaroglu20icbdr, xiang21fig, ji21jasa}.

\rev{Notice that CVMs cannot account for the internal compactness of each class in isolation, but the CVM of a class partition will get worse if some of these classes lack compactness or split across several clusters.} 

\subsection{Pitfalls}

\label{sec:cvmassumption}
The general process of label-based DR evaluation promotes embeddings with good CLM regardless of the CLM of the original data (\autoref{sec:intro}).
In other words, the process examines the extent to which CLM is harmed in embeddings while assuming that the original data has good CLM.
Thus, if the assumption is broken, the process will frame embeddings that correctly represent overlapped classes to have False Groups distortions. 
As the process considers good CLM embeddings as high-quality ones, it is also incapable of detecting Missing Groups distortions that may arise from CLM amplification. These pitfalls were identified for the first time by Aupetit \cite{aupetit14beliv}. \rev{Our preliminary experiment confirms such a threat (Appendix D). The general process of label-based evaluation erroneously prefers DR techniques that maximize the separation among classes, instead of the ones that aim to preserve the original structure of data if the datasets have bad CLM.}
Here, we aim to introduce a new way of using class labels for DR evaluation that mitigates such a bias.

\section{\LS \& \LC}

\label{sec:lsc}

We introduce two distortion measures---\LT and \LC (\lsc)---as an alternative way of using class labels for DR evaluation. 
Our measures examine how CLM differs in \textit{both} the original and embedded spaces where CVM is used to quantify CLM.
\lt and \lc capture the False and Missing Groups distortions, respectively. 
The measures are named after Trustworthiness and Continuity, two local distortion measures that focus on capturing False and Missing Neighbors \cite{venna06nn}.

\subsection{Design Rationale}

\label{sec:overcom}

\noindent
\textbf{Inputs, output, and hyperparameters} \ltc take (1) the high-dimensional data $\mathbf{X}$, (2) its DR embedding $\mathbf{Z}$, and (3) class labels $\mathbf{\mathbf{P}_L} = \{P_{L,1}, P_{L,2}, \cdots P_{L,k} \}$ as inputs. Both \lt and \lc output a number between 0 and 1; a higher value indicates lower distortions and a better embedding. For hyperparameters, a CVM $m$ with distance function $\delta$ is given. If $m$ is an EVM, we need to additionally select the clustering technique $C$ as a hyperparameter (\autoref{sec:cvmprocess}). The $m$ should assign higher scores to better clusterings and range from 0 to 1 (refer to \autoref{sec:req} for a detailed explanation about this requirement). 

\noindent
\textbf{Step 1. Measuring CLM in the original and embedded spaces}
We apply CVM to both the original and embedded spaces to examine CLM. Here, unlike the general process of label-based DR evaluation that applies CVM to all classes at once, we apply CVM to every \textit{pair} of classes, so that we can take account of the relationships of classes in more detail.
Formally, we construct the class-pairwise CLM matrices $M(\mathbf{X})$ and $M(\mathbf{Z})$, where the $(i, j)$-th cell of the matrices $M(\mathbf{S})_{i,j}$ ($\mathbf{S} \in \{\mathbf{X}, \mathbf{Z}\}$) is defined as:
\[
\left\{
\begin{array}{rcl} 
m(\{P_{L, i}, P_{L, j}\}, \mathbf{S}, \delta) & \text{if} & i \neq j \text{ and $m$ is an IVM} \\ 
m(C(P_{L, i} \cup P_{L, j}, \delta), \{P_{L, i}, P_{L, j}\}) & \text{if} & i \neq j \text{ and $m$ is an EVM} \\
0 & \text{if} & i=j 
\end{array}
\right..
\]

\noindent
\textbf{Step 2. Computing distortion matrices}
We construct a matrix $M^{*} = M(\mathbf{X}) - M(\mathbf{Z})$ representing CLM distortions. We then compute $M^{FG}$ and $M^{MG}$, where $M^{FG}_{i,j} = (M^{*}_{i,j}$ if $M^{*}_{i,j} >0$, else $0)$, and $M^{MG}_{i,j} = (-M^{*}_{i,j}$ if $M^{*}_{i,j} <0$, else $0)$. 
$M^{FG}$ and $M^{MG}$ abstract the CLM decrement (False Groups) and increment (Missing Groups), respectively.

\noindent
\textbf{Step 3. Aggregating distortions}
Finally, we average the \rev{upper-diagonal elements of} $M^{FG}$ and $M^{MG}$ into final scores:

\begin{itemize}
    \item[] \hspace{7mm} \textsc{\LT}: $1 - \mbox{avg}_{i, j} \rev{M^{FG}_{i > j}}$ \vspace{-2mm}
    \item[] \hspace{7mm} \textsc{\LC}: $1 - \mbox{avg}_{i, j} \rev{M^{MG}_{i > j}}$.
\end{itemize}
Note that we subtract the average from 1 to make embeddings with fewer distortions receive higher quality scores.

\subsection{Selecting CVM for \lsc}

We establish the requirements for CVM to get proper \ltc scores and present suitable CVM options.
In this section, we denote $m(\mathbf{P}, \mathbf{S}, \delta)$ as a CVM score with respect to $\mathbf{P}$, $\mathbf{S}$, and $\delta$.
If $m$ is an IVM $m_I$, we set $m(\mathbf{P}, \mathbf{S}, \delta) = m_I(\mathbf{P}, \mathbf{S}, \delta)$. If $m$ is an EVM $m_E$, we set $m(\mathbf{P}, \mathbf{S}, \delta) = m_E(C(\mathbf{S}, \delta), \mathbf{P})$ with $C$, the chosen clustering technique.


\subsubsection{Requirements}

\label{sec:req}

We set the first three requirements based on the following proposition: to be used for \ltc, a proper CVM should be comparable across $\mathbf{X}$ and $\mathbf{Z}$.
In other words, $m$ shall consider only \textit{how well the given partition is clustered in the given data} and be invariant to the characteristics that differ between $\mathbf{X}$ and $\mathbf{Z}$ but are not related to the cluster structure. 
For example, the scaling of the pairwise distances should not alter the score. Otherwise, the evaluation will be unreliable; for example, we can simply manipulate \ltc scores by scaling the original or embedded data while there is no change in the cluster structure.

Previous works \cite{jeon22arxiv2, bendavid08neurips} set axioms defining how a CVM can be independent of such features.
They require CVMs to be invariant to the change of scale, dimensionality, and the number of points and classes and to have a fixed range. 
As $\mathbf{X}$ and $\mathbf{Z}$ already share the number of points and classes, we require CVMs to ensure the other three axioms. 

The first axiom requires CVMs to be invariant against the scaling of distances between points, which can be inherently different in $\mathbf{X}$ and $\mathbf{Z}$:
\begin{itemize}[leftmargin=0pt]
    \item [] \textbf{Scale Invariance \cite{bendavid08neurips}} \textit{A CVM $m$ is scale invariant if $\forall$ partition $\mathbf{P}$, data $\mathbf{S}$, and distance function $\delta$, $m(\mathbf{P}, \mathbf{S}, \delta) = m(\mathbf{P}, \mathbf{S}, \alpha \delta)$ $\forall \alpha > 0$ (where $\alpha \delta$ is a distance function satisfying $\alpha \delta(x,y) = \alpha \cdot \delta(x,y)$, $\forall x, y \in \mathbf{S}$.). }
    \item [] \textbf{(R1)} A CVM should satisfy scale invariance.
\end{itemize}

The second axiom focuses on the effect of the data dimension on the distance $\delta$, due to the so-called curse of dimensionality \cite{bellman66science}. 
The growing dimensions increase the average of pairwise distances while the variances remain constant \cite{beyer99icdt, francois07tkde, lee11pcs}, thus the differences between distances become negligible. 
To be used for \ltc, CVM should be shift invariant \cite{lee11pcs, lee14cidm} to cancel the shift of the average distances due to the different dimensions of $\mathbf{X}$ and $\mathbf{Z}$.
\begin{itemize}[leftmargin=0pt]
    \item[]  \textbf{Shift Invariance \cite{jeon22arxiv2}} \textit{A CVM $m$ is shift invariant if $\forall \mathbf{P}, \mathbf{S}, \delta$, $m(\mathbf{P}, \mathbf{S}, \delta) = m(\mathbf{P}, \mathbf{S}, \delta+ \beta)$ $ \forall \beta > 0$ (where $\delta + \beta$ is a distance function satisfying $(\delta + \beta)(x,y) = \delta(x,y) + \beta$, $\forall x, y \in \mathbf{S}$).}
    \item[]  \textbf{(R2)} A  CVM should satisfy shift invariance.
\end{itemize}

The final axiom is about requiring CVMs to produce scores that conform to a fixed range of values, which is designed to capture the remaining subtle factors that are not influenced by the cluster structure. 
\begin{itemize}[leftmargin=0pt]
    \item[]  \textbf{Range Invariance \cite{jeon22arxiv2}} \textit{A CVM $m$ is range invariant if $\forall \mathbf{S}, \delta$, $\min_{\mathbf{P}} m(\mathbf{P},\mathbf{S},\delta)=\alpha$ and $\max_{\mathbf{P}} m(\mathbf{P},\mathbf{X},\delta)=\beta$, where $\alpha, \beta$ are constants satisfying $\alpha < \beta$ (arbitrarily set to 0 and 1, respectively).}
    \item[]  \textbf{(R3)} A CVM should satisfy range invariance.
\end{itemize}

Additionally, we want CVMs to be stable against the change of hyperparameters. 
This is because the alteration of CVM scores due to the hyperparameter change can induce uncertainty in utilizing \ltc. 
This leads to the last axiom:

\begin{itemize}[leftmargin=0pt]
    \item[]  \textbf{(R4)} A CVM should have no hyperparameter or should produce similar scores on the same input regardless of the hyperparameter settings.
\end{itemize}

\subsubsection{CVM Candidates}
\label{sec:lsc-examples}

We examine CVMs commonly used for DR evaluation (\autoref{sec:cvmprocess}) as potential candidates to be used for \ltc. 
\rev{For EVMs, we find that the combination of $K$-Means and adjusted rand index cannot be used. This is because the parameter $K$ (i.e., number of clusters) in $K$-Means
leads to the violation of \textbf{R4}. Indeed, as clustering techniques commonly require hyperparameters, EVMs hardly satisfy the aforementioned requirements. Studying how EVMs and clustering techniques can satisfy R4 is beyond the scope of this work.}

For IVMs,  neither the Silhouette Coefficient \cite{rousseuw87silhouette} nor the Davies-Bouldin index \cite{davies79tpami} satisfies shift invariance (\textbf{R2}; refer to Appendix A for the proof). However, we found that DSC satisfies all requirements, setting it as a proper CVM for \ltc (Appendix A).

We additionally found that the between-dataset Calinski-Harabasz index ($CH_{btwn}$) \cite{jeon22arxiv2}, a variant of Calinski-Harabasz index \cite{calinski74cis}, satisfies the four requirements: satisfaction of \textbf{R1}, \textbf{R2}, and \textbf{R3} has been demonstrated earlier \cite{jeon22arxiv2}; it also satisfies \textbf{R4} as its unique hyperparameter is the number of Monte-Carlo simulations for normalizing the measure, which barely affects the output if the number is sufficiently high.
We give a brief description of these two CVMs usable for \ltc:

\noindent
\textbf{Distance Consistency (DSC) \cite{sips09cgf}} DSC is defined as the number of data points closer to the centroid of another class than their own in the data, normalized by the total number of data points.  As DSC ranges from $0.5$ to $1$ \rev{if the number of classes is two} and assigns a lower score for a better CLM, we use the value $2(1 - \text{DSC})$ to make it satisfy \textbf{R3} (\autoref{sec:overcom} (Step 1)).

\noindent
\textbf{Between-dataset Calinski-Harabasz index ($CH_{btwn}$) \cite{jeon22arxiv2}}
$CH_{btwn}$ is defined as the ratio of compactness to separability. Compactness is defined as the distance between data points and the class centroids to which each point belongs, and separability is determined by the distances between class centroids and the centroid of the entire data.

\begin{figure}[t]
    \centering
    \includegraphics[width=\linewidth]{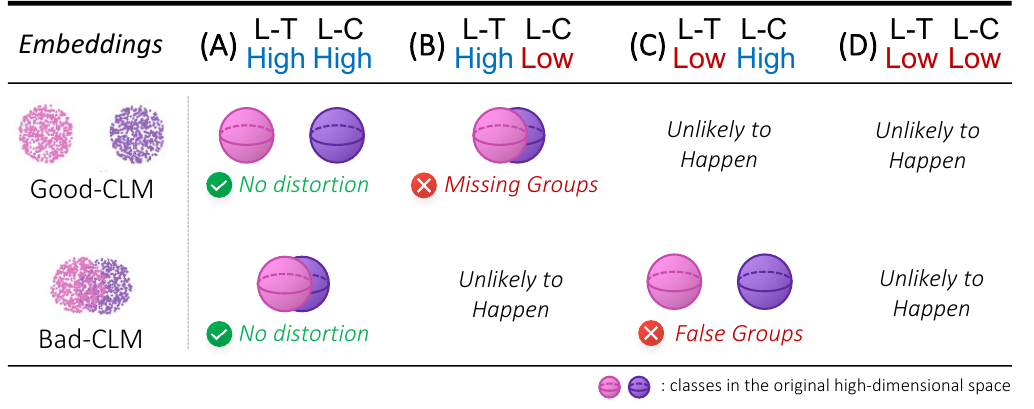}
    \vspace{-6mm}
    \caption{Guidelines to infer the CLM of the high-dimensional data based on the CLM of the embedded data (left column) and the scores given by \lt (L-T) and \lc (L-C) (first row) (see \autoref{sec:guideline} for details).
    }
    \vspace{-3mm}
    \label{fig:guide}
\end{figure}

\subsection{Guidelines to Interpret \ltc}

\label{sec:guideline}

We present the guidelines to interpret embeddings based on \ltc.
If \lt and \lc are both high,  the CLM of the embedding accurately depicts the CLM in the original space (\autoref{fig:guide}A). 
High \lt and low \lc (\autoref{fig:guide}B) mean that Missing Groups distortions have occurred, i.e., the CLM of the original data is worse than it appears in the embedding (first row); some pairs of classes appear more separated than they actually are in the data space. 
When the CLM of the embedding is already low (\textit{e.g.} due to overlapping classes), Missing Groups distortions are more unlikely to happen as the CLM in the data would have to be even worse (second row).
In contrast, low \lt and high \lc (\autoref{fig:guide}C) inform that False Groups distortions have occurred; the CLM in the original data is better than in the embedding (second row).
As False Groups distortions deteriorate the CLM of the embedding, the situation is unlikely to occur if the embedding has a good CLM, and thus can hardly become better (first row).
\rev{Due to such a tradeoff between False and Missing Groups (i.e., more Missing Groups lead to fewer False Groups, and vice versa), }
it is unlikely to get low \lt and \lc at the same time (\autoref{fig:guide}D). \rev{Our sensitivity analysis (\autoref{sec:seneval}; \autoref{fig:senexp}) confirms the existence of the tradeoff. }

\subsection{Time Complexity}

The complexity of \ltc depends on the CVM. 
As DSC is  $O(|\mathbf{S}||\mathbf{P}_L|\Delta_\mathbf{S})$, where  $\Delta_\mathbf{S}$ denotes the dimensionality of $\mathbf{S}$, applying it to a pair of classes $P_{L,i}, P_{L,j}$ requires $O(|P_{L,i} \cup P_{L,j}|\Delta_\mathbf{S})$. As each class is considered $|\mathbf{P}_L|$ times, \ltc with DSC is $O(|\mathbf{S}||\mathbf{P}_L|\Delta_\mathbf{S})$. 
Similarly, as \CHb is $O(|\mathbf{S}||\mathbf{P}_L|^2\Delta_\mathbf{S})$, applying it to a pair of classes $P_{L,i}, P_{L,j}$ requires $O(|P_{L, i} \cup P_{L, j}|\Delta_\mathbf{S})$. Therefore,  \ltc with \CHb is $O(|\mathbf{S}||\mathbf{P}_L|\Delta_\mathbf{S})$. In both cases, the time complexity is linear in all variables. We  evaluate the scalability of \ltc in \autoref{sec:scaleval}.

\subsection{Implementation \& Deployment}

We deploy \ltc as a Python library. We provided an interface that allows users to implement and test custom CVM as a hyperparameter. The source code is available at \href{https://github.com/hj-n/ltnc}{\texttt{github.com/hj-n/ltnc}}.

\begin{figure*}[ht]
    \centering
    \includegraphics[width=\textwidth]{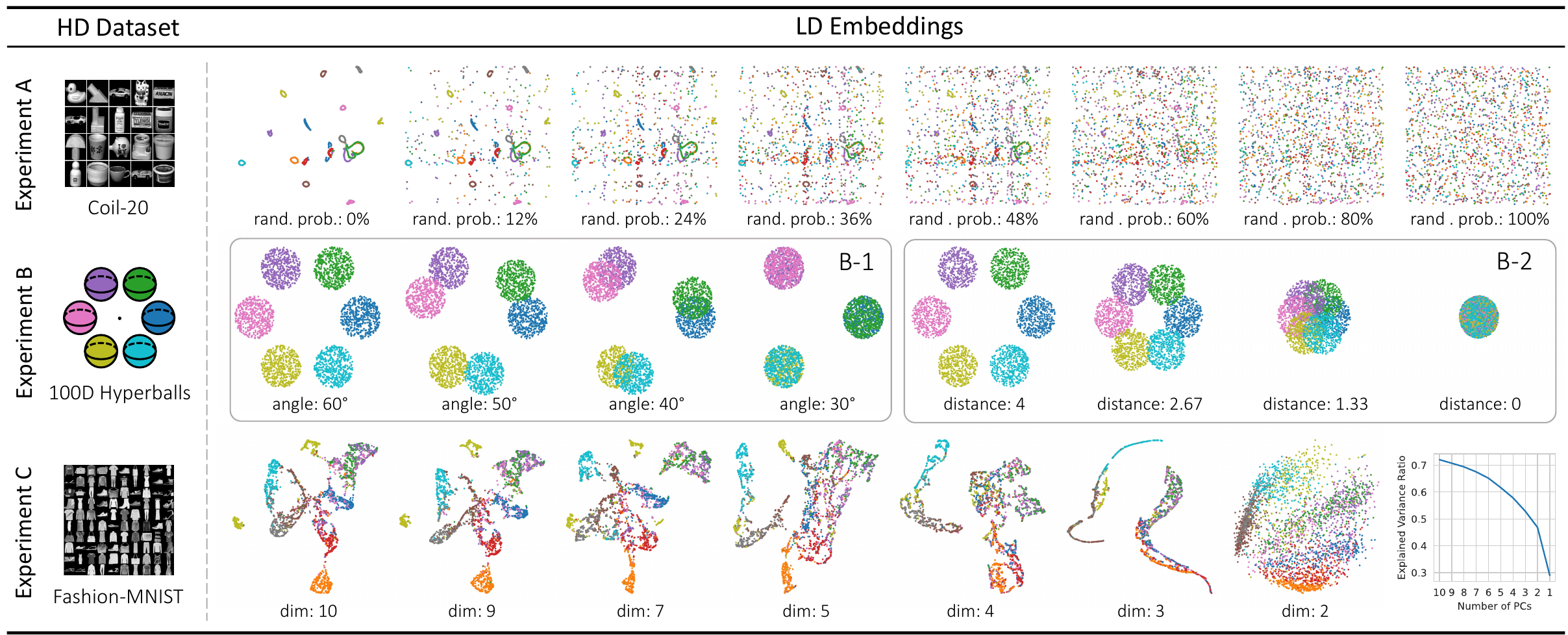} \vspace{-5.5mm}
    \caption{The high-dimensional (HD) datasets and low-dimensional (LD) embeddings used in experiments A, B, and C of sensitivity analysis (\autoref{sec:seneval}). The experiments aim to check the distortion measures' ability to capture False Groups distortions. Class labels are mapped to colors. (A) The Coil-20 \cite{nene96tech} dataset and the embeddings generated by randomizing the positions of the embedded points with a certain probability. 
    (B) A HD dataset consists of six well-separated hyperballs (left) and its synthetic embeddings (right) made by initializing the embedding with six well-separated discs and gradually overlapping the discs in two different manners (B-1, 2). (C) The Fashion-MNIST \cite{xiao2017arxiv} dataset and the PCA embeddings with different numbers of principal components (PC); here we depict the UMAP projection of PCA embeddings if it has more than two PCs (i.e., dimensionality is higher than two). We depict the relation between explained variance ratio and the number of PC in the line chart next to the embeddings. \vspace{-1.5mm}}
    \label{fig:sendataabc}
\end{figure*}

\begin{figure*}[ht!]
    \centering
    \includegraphics[width=\textwidth]{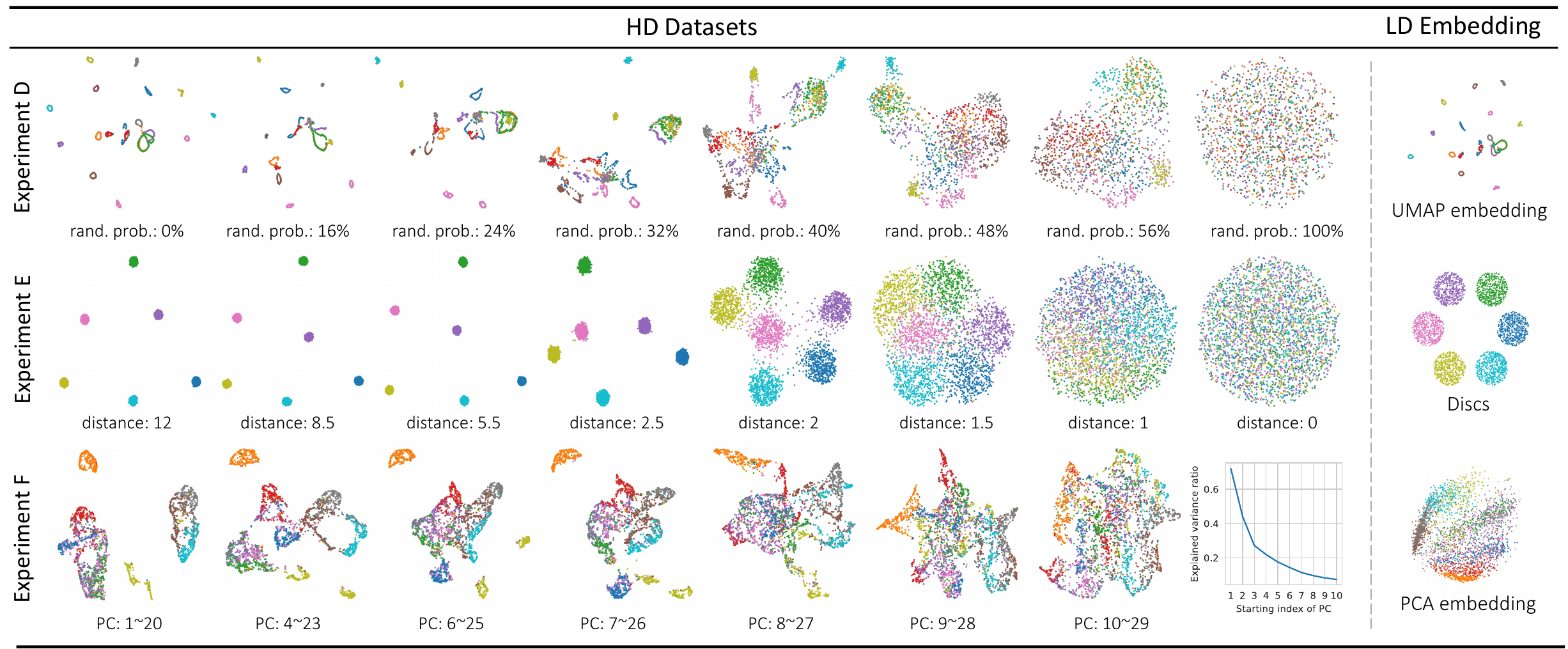} \vspace{-5.5mm}
    \caption{The low-dimensional (LD) embeddings and corresponding high-dimensional (HD) datasets represented as UMAP embeddings, used in experiments D, E, and F of sensitivity analysis 
    (\autoref{sec:seneval}) to examine distortion measures' ability to capture Missing Groups distortions.
    (D) An UMAP embedding of the Coil-20 \cite{nene96tech} dataset (right), and the variants of the Coil-20 dataset made by randomizing the coordinates of data points in HD space with a certain probability. 
    (E) A 2D embedding with six well-separated discs and synthetic HD datasets. We create the datasets by generating six 100D hyperballs and gradually overlapping them. (F) A 2D PCA embedding of the Fashion-MNIST dataset and corresponding HD datasets variants, created by slicing 20 principal components (PC) with different rankings. The line chart shows their corresponding explained variance ratio. \vspace{-2.5mm}}
    \label{fig:sendatadef}
\end{figure*}

\section{Quantitative Evaluations and Discussions}

We conduct quantitative experiments to evaluate \ltc with DSC and \CHb, i.e., \ltc [DSC] and \ltc [\CHb{}], respectively. In the sensitivity analysis  (\autoref{sec:seneval}), we check the accuracy of \ltc and competitors in quantifying distortions.
We also evaluate the runtime of the measures (\autoref{sec:scaleval}).

\noindent
\textbf{Competitors.} 
We first consider all distortion measures without labels (\autoref{sec:disme}) as competitors. 
For global measures, we use KL divergence and DTM. T\&C and MRRE are used as representative local measures. MRRE [Missing] and MRRE [False] target Missing and False Neighbors, respectively.
We select S\&C as the sole pair of measures targeting cluster-level distortions. 
For the measures using labels (\autoref{sec:dislabel}), we first add CA-T\&C. We then select Silhouette and DSC as representative CVMs used in the general label-based evaluation. 
For T\&C, MRRE, and CA-T\&C, we average their score across $k$-nearest neighbor values: $k = [5, 10, 15, 20, 25]$, following Jeon et al. \cite{jeon21tvcg}. For KL divergence and DTM, we average the scores across different standard deviation values of Gaussian kernels $\sigma$: $[0.01, 0.1, 1]$, following Moor et al. \cite{moor20icml}. For S\&C, we use the default hyperparameter setting \cite{jeon21tvcg}.

\subsection{Sensitivity Analysis}
\label{sec:seneval}

\label{sec:eval}
\begin{figure*}[ht]
    \centering
    \includegraphics[width=\textwidth]{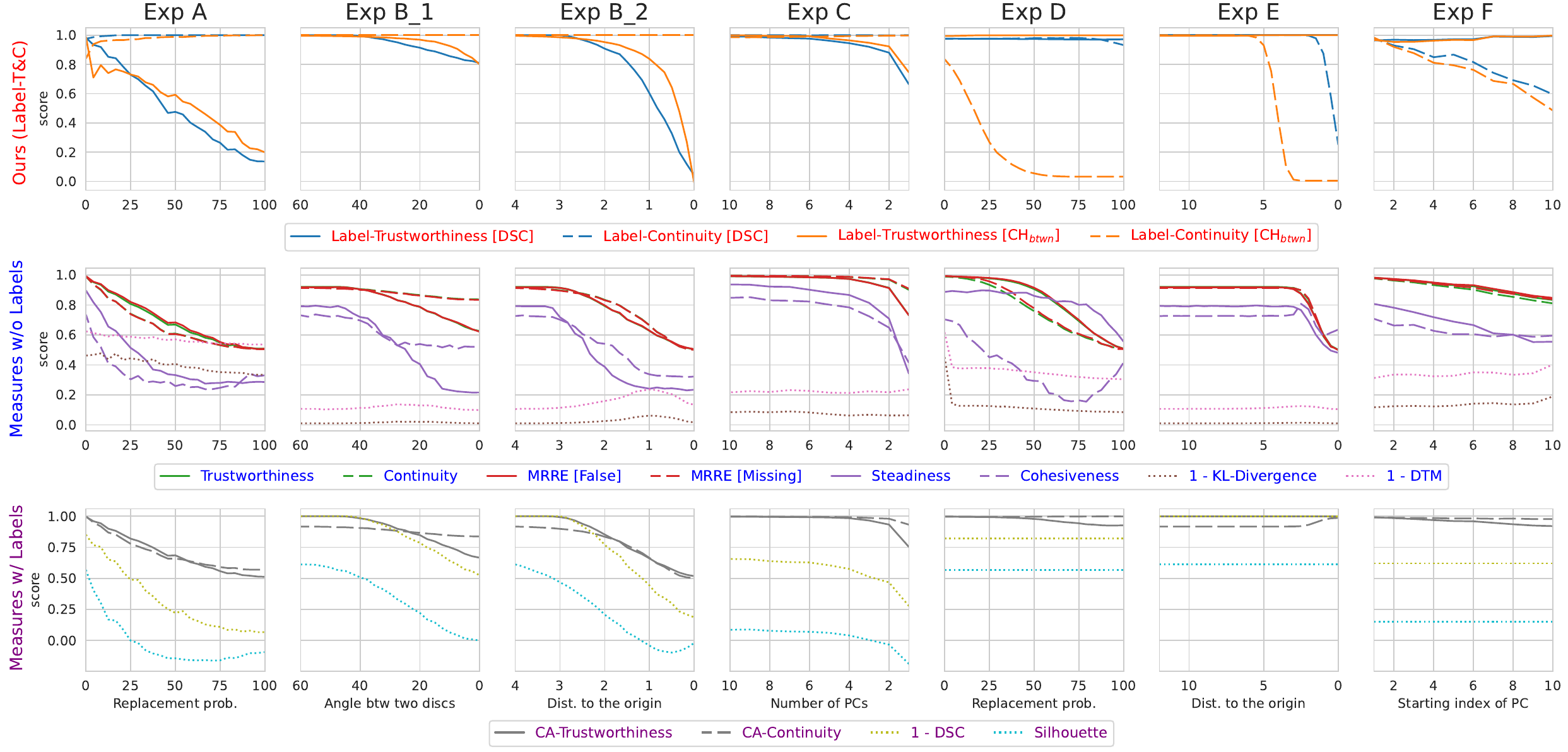} \vspace{-6mm}
    \caption{The results of the sensitivity analysis (\autoref{sec:seneval}; experiments A-F). Solid lines and dashed lines represent the measure that focuses on compression (e.g., False Groups, False Neighbors) and stretching (e.g., Missing Groups, Missing Neighbors), respectively. \rev{Dotted} lines represent global measures and CVMs. A pair of compression and stretching measures is represented with the same line color. 
    Measure names in red, blue, and purple correspond to our approach, the measures without labels (\autoref{sec:disme}), and the measures with labels (\autoref{sec:dislabel}), respectively.
    In summary, \lt (blue and orange bold line) and \lc (blue and orange dotted line) accurately detect Missing and False Groups distortions, respectively. Meanwhile, all other measures, including general label-based DR evaluation (i.e. DSC and Silhouette), fail to capture these distortions. \vspace{-3mm}}
    \label{fig:senexp}
\end{figure*}

We conduct six experiments (A-E) to examine \ltc{}'s sensitivity in quantifying False Groups (Fixed data and variable embeddings in experiments A, B, and C) or Missing Groups (Variable data and fixed embeddings in experiments D, E, and F) distortions.
The labeled data and embeddings used in the experiments can be found in \autoref{fig:sendataabc} (A, B, and C) and \autoref{fig:sendatadef} (D, E, and F).
In all of them, we run \ltc and competitors to evaluate the embeddings. 

\subsubsection{Objectives and Design}

\noindent
\textbf{Experiment A: Randomizing embeddings}
We examine whether \ltc and competitors can accurately quantify False Groups distortions. 
We generate a 2D UMAP embedding of the Coil-20 \cite{nene96tech}  dataset. We then create variants of the embedding with different levels of False Groups distortions by randomizing the location of the points.
We create 21 variants, ranging the replacement probability from 0\% (same as the original embedding) to 100\% (totally randomized) with an interval of 5\%. 
The original class assignments of Coil-20 are used as labels. 
We hypothesize that \lt will decrease as the replacement probability grows, properly capturing False Groups distortions, while \lc will ignore the distortions.

\noindent
\textbf{Experiment B: Overlapping discs}
We aim to check distortion measures' ability to precisely capture False Groups distortions, as with experiment A.
We create a high-dimensional dataset consisting of six hyperballs with a radius of 5 lying in 100 dimensions. 
We set the hyperballs to be equidistant ($=10$) from the origin. We then create an artificial 2D embedding consisting of six discs (radius of 1.5) evenly and equidistantly ($=4$) distributed around the origin $O$.
Data points and labels within each disc correspond to those of each hyperball.
The positions of each point within the disc and hyperball are determined randomly.
The label is also set based on the disc each point belongs to.
We gradually overlap the discs to artificially generate distortions. Here, we use two overlapping schemes to evaluate the sensitivity of \ltc in detail, resulting in two separate subexperiments (B-1, B-2). In B-1, three independent pairs of adjacent discs are overlapped; for each pair of discs $(A, B)$ with centers $C_A$, $C_B$, we adjusted $\angle C_A O C_B$ from $60^{\circ}$ to $0^{\circ}$ with an interval of $2.4^{\circ}$ (25 embedding variants in total). In B-2, we overlap all discs at once by moving them toward the origin; for each disc $A$, we gradually decrease $C_A O$ from 4 to 0 with an interval of 0.16 (25 embedding variants). 
We hypothesize that the \lt score will go down as False Groups distortions increase due to the overlap of the discs, while \lc will stay still. 
We also hypothesize that \lt will decrease more in B-2 than in B-1, as the overlap is larger.

\noindent
\textbf{Experiment C: Decreasing the dimension of the embedded space}
We generate False Groups distortions by decreasing the dimensionality of embedded space and check whether the measures can detect the distortions. 
We prepare the Fashion-MNIST \cite{xiao2017arxiv} as a high-dimensional dataset. We generate PCA embeddings with a decreasing number of top principal components (10 to 1 with an interval of 1; 10 embeddings in total). 
We expect the embeddings with a smaller number of principal components (i.e., embeddings lying in the space with fewer dimensions) to have more False Groups distortions as they have a smaller explained variance ratio (line chart in \autoref{fig:sendataabc}). 
We use the class assignments of the Fashion-MNIST dataset as labels.
Our hypothesis is that \lt will decrease as the dimensionality decreases, while \lc will stay still.

\noindent
\textbf{Experiment D: Randomizing the original data}
We want to evaluate \ltc and competitors' capability in accurately quantifying Missing Groups distortions. 
We first generate a fixed 2D UMAP embedding of the Coil-20 \cite{nene96tech} dataset. 
We then generate the variants of the original data by mixing the points in the high-dimensional space with a fixed probability, producing Missing Groups distortions. We control the replacement probability from 0\% to 100\% with an interval of 5\%, resulting in 21 variants. 
The class assignments of the original data are used as labels.
We hypothesize that \lc will decrease as Missing Groups distortions increase (i.e., replacement probability increase), and that \lt will ignore the distortions.

\noindent
\textbf{Experiment E: Overlapping hyperballs}
We want to evaluate whether \ltc and competitors can precisely capture Missing Groups distortions. 
We prepare variants of high-dimensional data and fixed low-dimensional embedding consisting of six 100D hyperballs and corresponding 2D discs, respectively. 
The points within the same disc have the same label. All discs are well separated from each other.
We artificially overlap hyperballs to generate Missing Groups distortions. 
For each hyperball $A_H$, we gradually decrease $C_{A_H}O$ from 4 to 0 with an interval of 0.16 (25 variants in total). 
We hypothesize that \lc will decrease as hyperballs overlap, while \lt will stay still.

\noindent
\textbf{Experiment F: Decreasing the dimension of the original data space}
We examine whether the distortion measures can detect the Missing Groups distortions made by the decrease in the dimensionality of the original data. We prepare a 2D PCA embedding of the Fashion-MNIST dataset. 
We then select ten 20D PCA embeddings with different sets of principal components as high-dimensional datasets; the $i$-th dataset variant consists of the $(i)$-th to $(i+19)$-th principal components, where $1 \leq i \leq 10$.
We expect the dataset with a higher order to have more Missing Groups distortions over the embedding as they have a smaller explained variance ratio (line chart in \autoref{fig:sendatadef}). 
We used the class assignments of the Fashion-MNIST dataset as labels.
We hypothesize that \lc will decrease as the starting index of principal components increases, while \lt will stay still.

\subsubsection{Results}

\autoref{fig:senexp} shows the results of our experiments that we comment on below. 

\noindent
\textbf{Experiment A} As the randomization probability grows, both \lt [DSC] and \lt [\CHb{}] similarly decrease linearly while \lc [DSC] and \lc [\CHb{}] slightly increase, confirming our hypothesis. 
Meanwhile, S\&C and local measures decrease regardless of the distortion type, while global measures slightly increase.
In the case of label-based measures, both CA-T\&C and the general CVM-based process (DSC and Silhouette) show mainly decreasing scores.

\noindent
\textbf{Experiment B} In B-1, as the overlap between the discs grows, both \lt [DSC] and \lt [\CHb{}] decrease in a similar manner, while \lc{}s stay still. 
Such results validate our hypothesis, confirming \ltc's capability in properly detecting False Groups distortions. 
Meanwhile, S\&C,  T\&C, and MRREs all decrease, while Steadiness, Trustworthiness, and MRRE [False] decrease more than Cohesiveness, CA-Continuity, and MRRE [Missing], respectively. 
Global measures stay still.
CA-T\&C partially succeed in properly detecting False Groups distortions; both CA-Continuity and CA-Trustworthiness decrease, but CA-Continuity's decrement was subtle compared to the one of CA-Trustworthiness. 
CVMs show a decreasing trend.
In B-2, the amount of decrement becomes bigger than in B-1 for \lt [DSC] and \lt [\CHb{}] while \lc{}s again stay still, confirming our second hypothesis.
The amount of decrement also becomes bigger than in B-1 for  T\&C, MRREs, and Cohesiveness, while Steadiness showed a similar drop as in B-1. 
In the case of KL divergence, DTM, and Silhouette, the patterns are almost identical to B-1 except that the scores rebound when the discs are nearly overlapped. 
The decrement becomes bigger also for CA-T\&C and DSC.

\noindent
\textbf{Experiment C} As the number of PCs decreases, \lt{}s decrease while \lc{}s stay still, validating our hypothesis. 
Global measures (KL divergence, DTM) stay still while all other measures decrease.

\noindent
\textbf{Experiment D} As we increase the randomization probability, both \lc [DSC] and \lc [\CHb{}] decrease, while \lt{}s stay still, verifying our hypothesis. However, while \lc [DSC] decreases right before the data are perfectly mixed, \lc [\CHb{}] decreases from the start. 
For local measures, both T\&C and MRREs decrease. 
Steadiness decreases, while Cohesiveness suddenly goes up after decreasing for a while.
Global (KL divergence, DTM) measures increase in general.
CA-Trustworthiness goes down while CA-Continuity stays still, and CVMs (DSC and Silhouette) stay still. 

\noindent
\textbf{Experiment E.} When the overlap between hyperballs increases, both \lc [DSC] and \lc [\CHb{}] decrease, while \lt{}s stay still, verifying our hypothesis. However, as in experiment D, \lc [DSC] and \lc [\CHb{}] decrease differently; while \lc [DSC] decreases right before the hyperballs perfectly overlap, \lc [\CHb{}] decreases before \lc [DSC] does. 
Meanwhile, local measures (T\&C, MRRE) decrease, while global measures (KL divergence, DTM) stay still.
Steadiness decreases while Cohesiveness temporarily pops up when Steadiness starts to decrease. 
CA-Trustworthiness maintains a maximum score while the CA-Continuity score increases before the perfect overlap of the hyperballs. CVMs stay still.

\noindent
\textbf{Experiment F.} The results confirm our hypothesis;  as the starting index of the PCs that we slice increases, both \lc [DSC] and \lc [\CHb{}] decrease while \lt{}s stay still. 
Local measures (T\&C, MRRE) decrease, and global measures (KL divergence, DTM) stay still.
S\&C decrease, while Steadiness decreases more than Cohesiveness.
CA-T\&C show a similar trend; CA-Trustworthiness decreases, while CA-Continuity decreases to a smaller extent. CVMs stay still.

\subsubsection{Discussions}

\noindent
\textbf{\ltc and competitors' capability in detecting cluster-level distortions.}
The results from experiments A-C confirm that \lt is sensitive to False Groups distortions, while \lc is not, as we intended. 
Moreover, the difference between the B-1 and B-2 results validates \lt{}'s accuracy at measuring the amount of False Groups distortions.
The results from experiments D-F, on the other hand, confirm that \lc accurately captures Missing Groups distortions, while \lt ignores them.

The results also validate that previous measures fail to accurately detect the distortions or to distinguish specific distortion types.
Global measures (KL Divergence, DTM) hardly discover distortions for all six experiments. 
Local measures (T\&C, MRRE) fail to pinpoint specific distortion types; all measures decrease regardless of the type of distortion they aim to measure.
Cluster-level measures (S\&C) fail to distinguish False Groups distortions in experiments A-C. 
For experiments D-F, the situation is even worse; Steadiness reacts more sensitively to Missing Groups distortions although it was originally designed to aim at False Groups distortions.
CA-T\&C succeed in pinpointing False Groups distortions for B-1, but fails to do so for the remaining experiments. 

The general process of label-based DR evaluation based on CVMs (DSC and Silhouette) succeeds in detecting the False Groups distortions in experiments A-C.
However, in experiments D-F, the process fails to detect Missing Groups distortions. Moreover, the process does not have a specific focus on distortion type and thus cannot explain whether the False or Missing Groups distortions occurred.
Such results confirm the threat of using the general label-based evaluation of DR in practice, providing clear evidence for adopting \ltc instead.

\noindent
\textbf{Effect of CVM choice on \ltc.}
\ltc{}s with two different CVMs (DSC or $CH_{btwn}$) show a consistent pattern in experiments A-C. However, they behave differently in experiments D and E; 
\lc [$CH_{btwn}$] starts decreasing for the lower level of generated CLM distortions than \lc [DSC]. This observation may be CVM-specific as DSC and \CHb use different schemes in examining how the classes are clustered. 
In \lc [DSC], the score only drops when classes overlap. 
Therefore, \lc [DSC]  is sensitive to Missing Groups distortions only if the \textit{overlapped} classes in the original space are more separated in the embedding. 
In contrast, $CH_{btwn}$ decreases as the proximity between classes increases, whether the classes overlap or not. Thus, when proximity increases, \lc [\CHb{}]  is more sensitive to Missing Groups distortions than \lc [DSC].
The results indicate that \CHb has a larger range of variation, being more sensitive to CLM than DSC, but it is less sensitive to class overlap. 
Creating a CVM both sensitive to CLM and class overlap while fulfilling our requirements (\autoref{sec:req}) constitutes an interesting future work. 




\noindent
\textbf{Discussions on the competitors.}
We discuss the patterns shown by competitors with more detail in Appendix B.

\rev{

\subsubsection{Sensitivity Analysis with the Class Labels Generated by Clustering Techniques}

We want to validate whether the results of our study are replicable with the labels that come from other sources. We thus conduct experiments A-F while generating class labels with clustering techniques (Appendix F). We find that \ltc show consistent results regardless of the sources of labels, while the general label-based DR evaluation process (i.e., CVMs) fails to do so. Such results confirm the robustness of the \ltc in evaluating the quality of DR embeddings.

}


\begin{figure}[t]
    \centering
    \includegraphics[width=\linewidth]{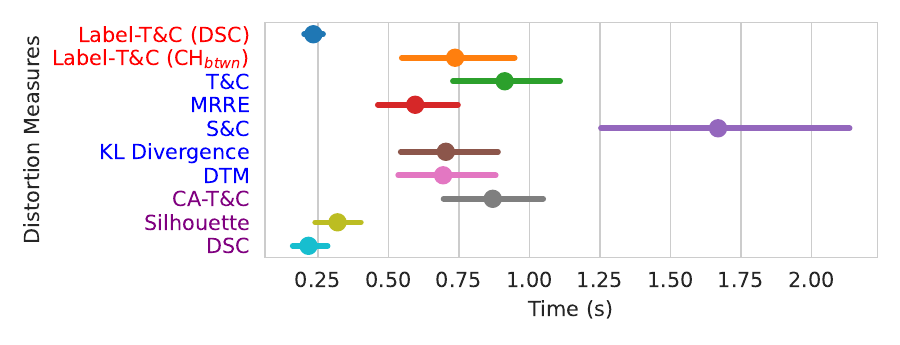}
    \vspace{-8.5mm}
    \caption{Results of the scalability analysis. Name and line colors match with \autoref{fig:senexp}. \ltc [DSC] (dark blue) is on par with CVMs (Silhouette, DSC), while \ltc [\CHb] is similar to most of the other measures. S\&C is the slowest. \vspace{-4mm}}
    \label{fig:scal}
\end{figure}

\begin{figure*}[t]
    \centering
    \includegraphics[width=\textwidth]{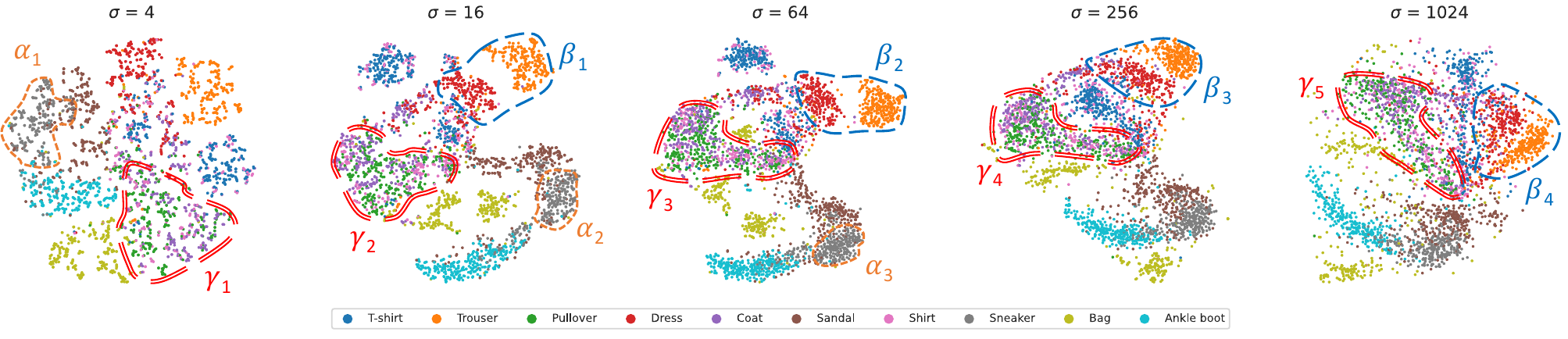} \vspace{-6mm}
    \caption{$t$-SNE embeddings of Fashion-MNIST \cite{xiao2017arxiv} data with diverse perplexity ($\sigma$) values. Combined with the class-pairwise CLM of the original dataset (\autoref{fig:app_pp_heatmap}), the patterns in the embeddings qualitatively support the findings about the effect of $\sigma$ revealed by \ltc (\autoref{fig:app_hp}; \autoref{sec:apptsne}). \vspace{-2mm}}
    \label{fig:app_pp}
\end{figure*}

\begin{figure}[t]
    \centering
    \includegraphics[width=\linewidth]{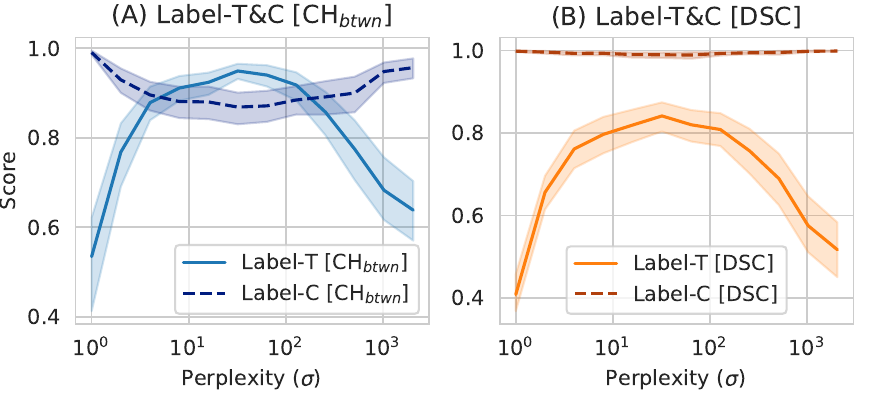} \vspace{-6mm}
    \caption{Overall reliability of $t$-SNE embeddings according to the $\sigma$ value quantified by \ltc [DSC] and \ltc [\CHb{}]. For each $\sigma$ value, we average the score of the embeddings generated from 94 labeled datasets (95\% confidence interval shaded).  \vspace{-2mm}}
    \label{fig:app_hp}
\end{figure}

\subsection{Scalability Analysis}

\label{sec:scaleval}

\subsubsection{Objectives and Design}

We evaluate the scalability of \ltc against the competitors. 
We gather 96 labeled datasets \cite{jeon22arxiv2} that vary in dimensionality,  the number of data points, and the number of classes. We exclude two datasets as the implementation of S\&C provided by the authors\footnote{\href{https://github.com/hj-n/steadiness-cohesiveness}{github.com/hj-n/steadiness-cohesiveness}} fails to process them, resulting in 94 datasets (Appendix C).
We generate embeddings using $t$-SNE, UMAP, PCA, and random projection for all 94 datasets.
We check the overall execution time applying all measures to the embeddings, adding up the running times of the measures run in pairs (\ltc, T\&C, MRRE, S\&C, and CA-T\&C). We use the provided implementation for S\&C and scikit-learn \cite{pedregosa11jmlr} for the Silhouette.  We implement the remaining measures in Python with Numba parallel computing \cite{lam15llvm} to maximize the scalability.
We run the experiments on a Linux server with 40-core Intel Xeon Silver 4210 CPUs.

\subsubsection{Results and Discussion}

\autoref{fig:scal} show that the running time of \ltc highly depends on the CVM. Among all measures, DSC is the fastest, followed by \ltc [DSC]. If $CH_{btwn}$ is used as the CVM, \ltc becomes less scalable. Still, \ltc [$CH_{btwn}$] has scalability similar to local (T\&C, MRRE) and global (KL Divergence, DTM) measures and to CA-T\&C,  all being more than twice faster than S\&C.

\begin{figure}[t]
    \centering
    \includegraphics[width=\linewidth]{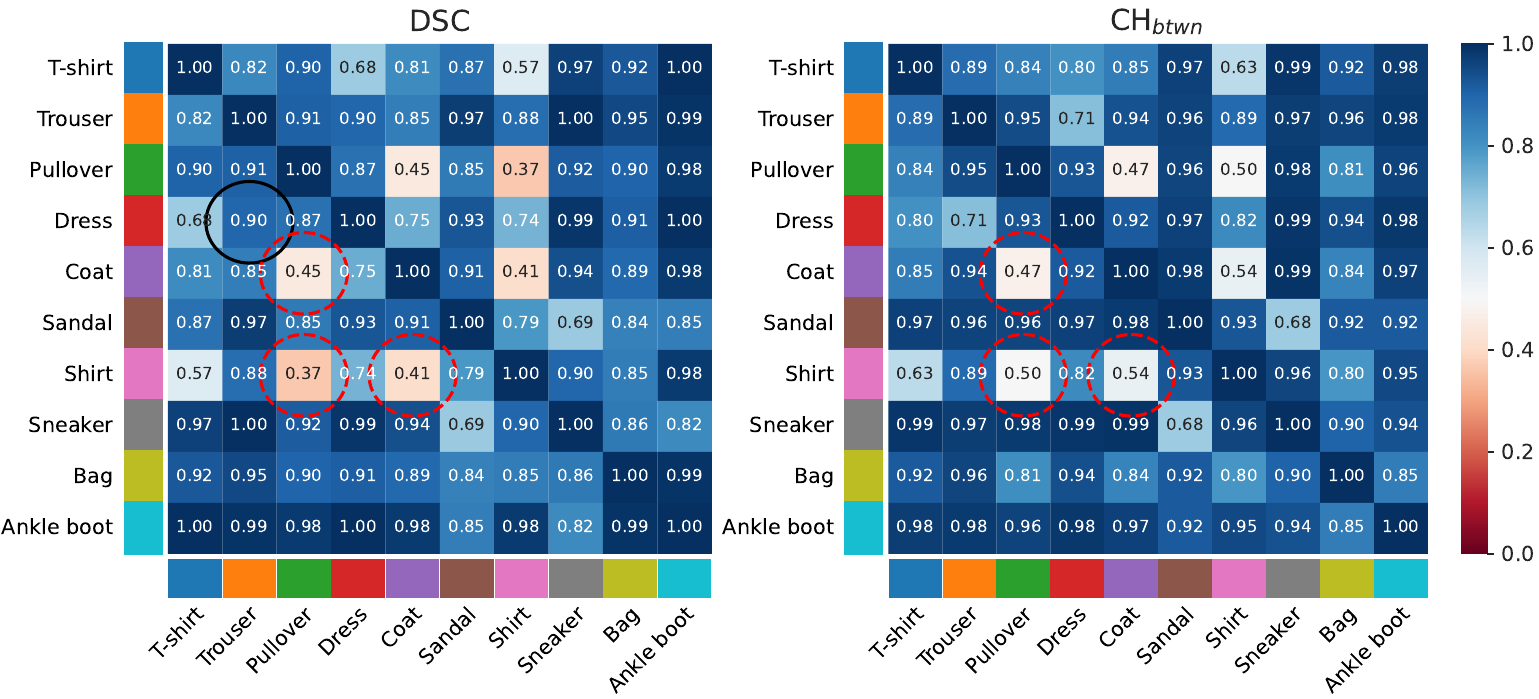} \vspace{-6mm}
    \caption{Heatmaps detailing the CLM matrix of the Fashion-MNIST dataset ($M(X)$ in \autoref{sec:overcom}). The color of each cell depicts the CVM (DSC, \CHb{}) score measured for each pair of classes corresponding to rows and columns. \vspace{-2mm}}
    \label{fig:app_pp_heatmap}
\end{figure}

\begin{figure*}[t]
    \centering
    \includegraphics[width=\textwidth]{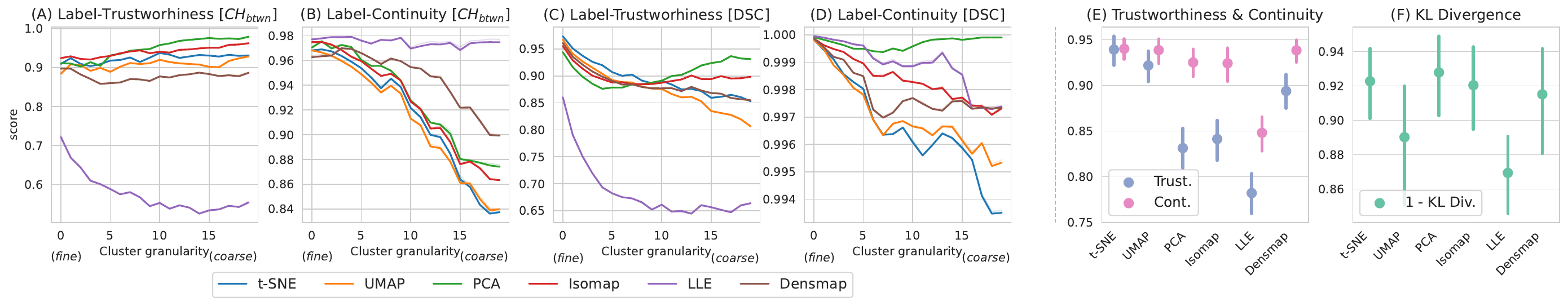}
    \vspace{-6mm}
    \caption{
    CLM distortion evaluation of a linear (PCA) and five nonlinear (t-SNE, UMAP, Isomap, LLE, and Densmap) unsupervised DR techniques. \rev{(A-D) Evaluation results with \ltc [\CHb{}/DSC] where} class labels are obtained from the hierarchical clustering of the original data at multiple granularity levels (x-axis). \ltc evaluates more coarse-grained (global) clusterings for higher levels. See details in \autoref{sec:app_hier}. 
    \rev{(E-F) Evaluation results of the techniques with T\&C (E) and KL Divergence (F). Note that for all figures, higher scores indicate better embeddings.} 
    \vspace{-3mm}}
    \label{fig:app_hier}
\end{figure*}

\section{Case Studies}

We report two case studies demonstrating the usefulness of \ltc to characterize DR techniques and their hyperparameters.

\subsection{Examining the Effect of \textbf{\textit{t}}-SNE Perplexity}

\label{sec:apptsne}

\subsubsection{Objectives and Design}

We want to use \ltc to evaluate the reliability of the cluster structures from $t$-SNE embeddings (\autoref{sec:seneval}) depending on its perplexity hyperparameter $\sigma$. $\sigma$ adjusts the balance between local and global cluster structures \cite{wattenberg2016tsnetuning, cao17arxiv}. 
We generate the $t$-SNE embeddings of the 94 labeled datasets used for the scalability analysis   (\autoref{sec:scaleval})  using different $\sigma$ values ($\sigma \in \{2^i \mid i=0,\cdots, 10 \}$) and evaluate them using \ltc [\CHb{}] and \ltc [DSC].
We also inspect the $t$-SNE embeddings of the Fashion-MNIST \cite{xiao2017arxiv} dataset with various perplexity values ($\sigma \in \{4, 16, 64, 256, 1024\}$; \autoref{fig:app_pp}) to gain more qualitative insights.
Moreover, we compute the ``ground-truth'' CLM matrix of the Fashion-MNIST dataset (\autoref{fig:app_pp_heatmap}), where the $(i, j)$-th cell represents the CVM score (\CHb or DSC) of the $i$-th and $j$-th classes. Note that this CLM matrix is identical to $M(\mathbf{X})$ in \autoref{sec:overcom}. 

\subsubsection{Results and Discussions}

\newcommand{\greytext}[1]{\textcolor{gray}{#1}}
\newcommand{\orangetext}[1]{\textcolor{orange}{#1}}
\newcommand{\redtext}[1]{\textcolor{Red}{#1}}
\newcommand{\greentext}[1]{\textcolor{ForestGreen}{#1}}
\newcommand{\purpletext}[1]{\textcolor{Plum}{#1}}
\newcommand{\pinktext}[1]{\textcolor{VioletRed}{#1}}


In the case of \ltc [\CHb{}], we found a clear tradeoff between \lt and \lc (\autoref{fig:app_hp}A).
When $\sigma$ is low or high, \lt [\CHb{}] gives low scores to $t$-SNE embeddings, indicating more False Groups distortions, while \lc [\CHb{}] gives high scores, meaning fewer Missing Groups distortions. 
This means that $t$-SNE underrepresents the extent to which classes are clustered. 
In contrast, when $\sigma$ has an intermediate value, \ltc [\CHb{}] indicate more Missing Groups and fewer False Groups distortions; hence, $t$-SNE exaggerates the degree to which classes are clustered. 

These results align well with the intent of $\sigma$. With low $\sigma$, $t$-SNE focuses more on a small number of neighbors, likely fewer than the clusters' sizes, interpreting each cluster as made of loosely-connected components in the data space. Thus, the embedding is more likely to split classes into several clusters in the embedding.
This phenomenon occurs in the Fashion-MNIST embedding (\autoref{fig:app_pp}); the \greytext{\textit{Sneaker}} class is less dense if $\sigma$ is low (\rev{region $\alpha_1$}) and relatively condensed when $\sigma$ has intermediate values (\rev{$\alpha_2$ and $\alpha_3$}).
For the latter, the number of neighbors that $t$-SNE focuses on will likely match the size of natural clusters within the original data. Therefore, $t$-SNE embeddings will tend to dismiss the inter-cluster connections, exaggerating the between-cluster distances. The number of neighbors that $t$-SNE focuses on with high $\sigma$ values will likely be bigger than the clusters' sizes. Thus, $t$-SNE will detect all data clusters as one densely-packed component and generate embeddings with smaller inter-cluster distances.

The relation between the \orangetext{\textit{Trouser}} and \redtext{\textit{Dress}} classes of the Fashion-MNIST embeddings (\autoref{fig:app_pp}) qualitatively verifies these hypotheses. Their DSC scores are almost maximum (the black circle in \autoref{fig:app_pp_heatmap}), meaning they slightly overlap in the data space. However, their distance in the embedding is exaggerated with intermediate $\sigma$ (\rev{$\beta_1$ and $\beta_2$}) compared to high $\sigma$ (\rev{$\beta_3$ and $\beta_4$}).
The same effect was observed qualitatively by Jeon et al. \cite{jeon21tvcg} while \ltc does so quantitatively.

Meanwhile, \lc [DSC] decreases slightly for intermediate values of $\sigma$ (\autoref{fig:app_hp}B dotted line). 
As \ltc [DSC] focuses more on class overlaps and less on between-class distances compared to \ltc [\CHb{}] (see D and E in \autoref{sec:eval}), it indicates that $t$-SNE preserves well the extent to which classes overlap regardless of $\sigma$.
To quantitatively validate these findings, we searched for the overlapped classes within the CLM matrices, assuming that $t$-SNE accurately depicts class overlap for all $\sigma$ values. We observed that the \greentext{\textit{Pullover}}, \purpletext{\textit{Coat}}, and \pinktext{\textit{Shirt}} classes overlap in the high-dimensional space (red circles in \autoref{fig:app_pp_heatmap}; both their DSC and \CHb class-pairwise scores are low). 
We found that these classes overlap in all embeddings in \autoref{fig:app_pp} (\rev{$\gamma_1$ to $\gamma_5$}), confirming our assumption.

In summary, we can conclude that for non-overlapping classes in $t$-SNE embeddings, the amount of proximity between them depends essentially on $\sigma$ and is not indicative of the proximity of these classes in the data space: $t$-SNE is not trustworthy regarding the original distance between visually separated classes.  However, classes with strong overlaps in the data are depicted as overlapping in the embedding too: $t$-SNE is more trustworthy for overlapping classes. 
Such results align with the qualitative findings of Wattenberg et al. \cite{wattenberg2016tsnetuning}. 

Overall, these findings demonstrate the effectiveness of \ltc to enhance our understanding of the effect of $\sigma$ on $t$-SNE results. 
We conduct the same analysis utilizing the competitor measures we used in our evaluation (\autoref{sec:eval}); refer to Appendix E for the results.

\subsection{Analyzing DR Techniques' Performance in Detail}

\label{sec:app_hier}

\subsubsection{Objectives and Design}

We use \ltc to analyze the quality of unsupervised DR techniques across fine-grained to coarse-grained cluster structures. 
We embed each of the previous 94 datasets using six DR techniques: $t$-SNE, PCA, UMAP, Isomap, LLE, and Densmap \cite{narayan21nature}. 
We also apply hierarchical clustering, getting 20 clustering partitions with different granularity levels for each of these datasets. 
The levels of granularity are obtained by thresholding the pairwise distances computed by Ward linkage \cite{ward64taylor} into 20 equal ranges.   
We use \ltc [\CHb{}] and \ltc [DSC] to evaluate the embeddings using each of the 20 clusterings as class labels.

\rev{
We also want to check whether the results obtained by \ltc align with the ones made by previous measures. 
We thus evaluate the embeddings using T\&C and KL divergence as representative local and global measures, respectively. We use the same hyperparameter setting with the sensitivity analysis (\autoref{sec:seneval}).
}

\subsubsection{Results and Discussions}
\autoref{fig:app_hier} depicts the results. 
LLE generates few Missing Groups distortions (highest \lc score; \autoref{fig:app_hier}B, D) at any level, but more False Groups distortions as the granularity level increases (\lt decreases; \autoref{fig:app_hier}A, C). \rev{This finding aligns with the fact that LLE obtains the worst KL divergence score among all techniques (\autoref{fig:app_hier}F).}
Such results are coherent with how LLE works, trying to reconstruct the ``local patches'' consisting of each point and its nearest neighbors while neglecting the overlap between the patches. 

There is a \lc downward trend across all other techniques as the level increases,
while \lc [DSC] shows higher scores than \lc [\CHb{}] (\autoref{fig:app_hier}B, D). 
This implies that Missing Groups distortions generally occur more for coarse-grained structures than for fine-grained ones; DR techniques exaggerate the separation between clusters at a global level.
$t$-SNE and UMAP especially give the worst  \lc scores because they focus on the preservation of local neighborhoods, casting doubts on their reliability in identifying global clusters. \rev{T\&C and KL divergence score provide strong evidence to the reliability of that claim. $t$-SNE and UMAP are in the top-2 highest ranks for T\&C but fail to do so for KL divergence. }

For \ltc except \lc [\CHb{}], PCA gets the best score at higher  granularity, suggesting that PCA is more reliable to conduct global tasks such as the density and similarity identification of clusters.
\rev{These results align with the fact that PCA earns the best score for KL divergence.}
The phenomenon confirms the experimental observation made by Xia et al. \cite{xia22tvcg}. This is also coherent with the fact that PCA embeds the data along the top two principal axes that preserve most of their variance, better representing coarse-grained structures than fine-grained ones.

We also find that Densmap, which is a variant of UMAP better preserving cluster density \cite{narayan21nature}, gets worse \lt [\CHb{}] scores than UMAP (\autoref{fig:app_hier}A) but better \lc [\CHb{}] scores (\autoref{fig:app_hier}B), at all levels.
This means that Densmap generates fewer Missing Groups but more False Groups distortions than UMAP.
As Densmap approximately maintains the cluster locations of UMAP \cite{narayan21nature}, such difference indicates that the clusters generally become bigger in Densmap compared to UMAP, hence the cluster density is relatively lower. 
Meanwhile, Densmap gets better \ltc [DSC] scores than UMAP for high granularity levels, confirming Densmap's advantage in investigating the overlap of clusters. \rev{The result is consistent with the KL divergence scores, indicating Densmap's advantage in preserving global structures when compared to UMAP (\autoref{fig:app_hier}F). }

These findings confirm the ability of \ltc to reveal the characteristics of DR methods over a wide range of clustering granularities.
\rev{
Although typical evaluation approaches of DR quality using both local and global measures (\autoref{fig:app_hier}E, F) \cite{jeon22vis, moor20icml, espadoto21tvcg} show consistent results, they 
} cannot reveal how the quality changes across granularity levels, as different measures are incomparable.

\section{Conclusions}

The general process of label-based DR evaluation relies on the assumption that the original data has good CLM, which can lead to erroneous conclusions when this assumption is violated.
We introduce two new distortion measures---\LT and \LC  (\ltc)---that use class labels for DR evaluation while eliminating the need to check the validity of the CLM assumption. Our quantitative experiments show that \ltc outperforms previous DR measures in terms of precision and sensitivity in detecting Missing and False Groups distortions. Use cases show that \ltc can be used to characterize DR techniques and their hyperparameters.

As future work, we will study new CVM to make \ltc more sensitive to the CLM distortions than using DSC or \CHb. 
Enriching the embedding with CLM distortions \cite{lespinats11cgf} could also better inform analysts about the credibility of visual patterns. Yet another direction would be to evaluate supervised DR techniques with \ltc. 
\rev{We also believe that supervised DR techniques using class labels in their optimization process could benefit from incorporating \ltc in their loss function.}
Overall, our proposal aims toward getting more trustworthy DR-based visual analysis.

\acknowledgments{
This work was supported by NAVER Corporation (Cloud Data Box) and by the National Research Foundation of Korea (NRF) grant funded by the Korea government (MSIT) (No. 2023R1A2C200520911).
The authors thank Seokhyeon Park for his valuable feedback in improving the figures. The authors also appreciate Sungbok Shin and SNU HCIL members for their feedback in improving the language. }

\bibliographystyle{abbrv-doi}

\bibliography{ref}

\end{document}